\documentclass[10pt,letterpaper]{article}

\usepackage{cogsci}

\cogscifinalcopy 

\usepackage{pslatex}
\usepackage{apacite}
\usepackage{float} 

\usepackage{graphicx}
\usepackage{bm}
\usepackage{amsmath}
\usepackage[table, dvipsnames]{xcolor}
\usepackage{multirow}
\usepackage[dvipsnames]{xcolor}

\title{Cobweb: An Incremental and Hierarchical Model of\\ Human-Like Category Learning}

\author{{\large \bf Xin Lian (xinlian@gatech.edu)}\\
  Georgia Institute of Technology\\
  Atlanta, GA 30308 USA
  \And {\large \bf Sashank Varma (varma@gatech.edu)} \\
  Georgia Institute of Technology \\
  Atlanta, GA 30308 USA
  \AND {\large \bf Christopher J. MacLellan (cmaclell@gatech.edu)} \\
  Georgia Institute of Technology \\
  Atlanta, GA 30308 USA
  }

\begin{document}

\maketitle

\begin{abstract}

\textit{Cobweb}, a human-like category learning system, differs from most cognitive science models in incrementally constructing hierarchically organized tree-like structures guided by the category utility measure. Prior studies have shown that Cobweb can capture psychological effects such as basic-level, typicality, and fan effects. However, a broader evaluation of Cobweb as a model of human categorization remains lacking. The current study addresses this gap. It establishes Cobweb's alignment with classical human category learning effects. It also explores Cobweb's flexibility to exhibit both exemplar- and prototype-like learning within a single framework. These findings set the stage for further research on Cobweb as a robust model of human category learning.

\textbf{Keywords:} 
categorization, concept learning, prototypes, exemplars
\end{abstract}

\section{Introduction}



Learning a \textit{category} (or \textit{concept}) involves inferring its structure from a set of examples \cite{griffiths2008categorization}. Various computational models of concept learning have been proposed under a number of theoretical frameworks. Some are \textit{rule-based}, suggesting that concepts are represented as rules formulated within a compositional representation language \cite{kemp2012exploring}. These include RULEX \cite{nosofsky1994rule} which generates conjunctive rules and retains their exceptions, the Bayesian description of rules \cite{goodman2008rational}, integrated mental models \cite{goodwin2011mental}, and the algebra of concept learning \cite{feldman2006algebra}. Other approaches are \textit{similarity-based}, including exemplar- and prototype-based models. Representative examples are ALCOVE \cite{kruschke1992alcove} and SUSTAIN \cite{love2004sustain}, both of which are exemplar models representing concepts with the weights of connectionist networks.

Within similarity-based models, a subset involves models that employ rational analysis to learn concepts \cite{anderson1990rational, nosofsky1998optimal, ashby1995categorization, rosseel2002mixture}. These models support \textit{incremental learning and updating} of acquired knowledge. A limitation of many rational categorization models \cite{anderson1990rational, sanborn2006more, griffiths2007unifying} is that they predominantly propose flat partitions.
Such representations might not fully capture important psychological effects such as the typicality effect, specifically the processing of atypical instances. 

It is therefore interesting to consider \textit{Cobweb} \cite{fisher1987knowledge}, which learns concepts incrementally and hierarchically, organizing cognitive structures into hierarchical levels of partitions, making it a potentially powerful model of human-like category learning \cite{langley2022computational}. Cobweb has a long history in artificial intelligence and is noteworthy for its incremental learning capabilities. \citeA{fisher1990structure} demonstrates its ability to explain various psychological effects including basic-level, typicality, and fan effects. However, beyond these initial efforts, Cobweb remains underexplored as a model of human categorization.

In this paper, we further evaluate Cobweb's potential as a model of human category learning. We assess the alignment between predictions made from two different levels of its hierarchy (subordinate leaves and basic concepts) and the classical findings of \citeA{medin1978context} and \citeA{shepard1961learning}. Our work shows Cobweb's efficacy in accounting for human categorizations at a general level. Importantly, we observe that Cobweb does not rigidly adhere to prototype- or exemplar-like behavior. This flexibility arises from its hierarchical cognitive structure, which enables the generation of predictions that range from prototype- to exemplar-like. We demonstrate Cobweb's proficiency as an incremental learner and thus its ability to account for human category learning over multiple training blocks. 
Finally, we illustrate Cobweb's robust alignment with human categorization across various tasks. The versatility displayed by Cobweb underscores its potential as a comprehensive model of human categorization.

\section{Cobweb: Human-Like Categorization}


\begin{figure*}[t!]
\centering
\includegraphics[width=1.0\textwidth]{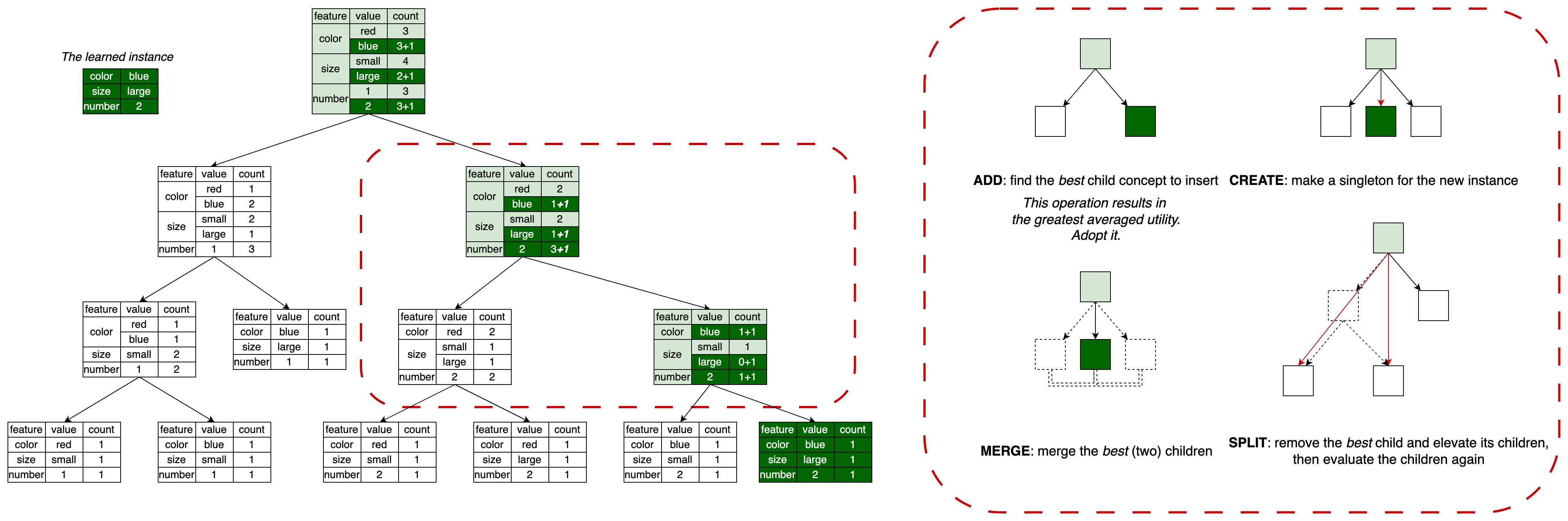}
\caption{An illustrative example of Cobweb's learning process which involves incorporating a new instance, depicted as a dark green table, into the existing tree structure. Cobweb traverses the tree from the root to a specific leaf node, and along this path, the concept nodes (highlighted in light green) are updated to reflect the given instance. The changes resulting from fitting the new instance into the tree are denoted in bold and italics. During this traversal, Cobweb considers four available operations at each branch, \textit{adding}, \textit{creating}, \textit{merging}, and \textit{splitting}. It then proceeds with the operation that yields the highest averaged category utility within the subtree. For instance, for the branch in the red dot box, Cobweb chooses to add the instance to the ``best'' child because it results in the highest average utility score.}
\label{fig:cobweb_learning}
\end{figure*}

Cobweb \cite{fisher1987knowledge} takes an incremental and hierarchical approach to learning. The system forms concepts given sequentially presented instances, which are represented as discrete attribute-value pairs
(e.g., \texttt{color: blue; form: triangle; size: large; number: 2}).
Given such examples, Cobweb forms a probabilistic concept hierarchy.
Each concept node in the hierarchy stores a probability table that tracks the frequency of each attribute value occurring in instances of the concept.
The left panel of Figure~\ref{fig:cobweb_learning} shows examples of Cobweb's instance and concept representations.
To guide the concept formation process, Cobweb uses \textit{Category Utility} ($CU$), which was proposed by \citeA{corter1992explaining} to account for human categorization effects.
This measure, which we define below, evaluates the feature predictive power of a concept $C_k$.


When a new instance $\bm{x}$ is introduced to be learned, Cobweb sorts it down its current categorization tree. 
At each node, Cobweb considers how best to incorporate the instance into the current node's children $\{C_k\}$. It evaluates four operations - \textit{add}, \textit{create}, \textit{merge}, or \textit{split} - and chooses the one that produces the highest averaged category utility:
\begin{equation}
    \frac{\sum_{k=1}^s CU(C_k)}{s}
\end{equation}
where $s$ is the number of children at the current branch. The average category utility measure lets Cobweb compare partitions with varying numbers of concepts  \cite{fisher1987knowledge}.

To evaluate the operations, Cobweb starts by simulating \textit{adding} the instance into each of the children concepts.
When an instance is added to a concept, its probability table frequencies are updated to reflect the instance's attribute values.
Cobweb uses these probability tables to efficiently compute the average category utility score without having to iterate over prior instances.
After considering each addition, Cobweb identifies the two children $C^1_k$ and $C^2_k$ that yield the highest and second-highest average category utility.

Next, the system evaluates \textit{merging}, \textit{splitting}, and \textit{creating}. To evaluate merging, Cobweb simulates the creation of a new child concept that merges the probability table counts of $C^1_k$ and $C^2_k$ (these concepts become children of the new concept) and updates the result to reflect the addition of the instance.
To evaluate splitting, it simulates removing $C^1_k$ and promoting its children to the current level. Finally, Cobweb considers creating a new concept that reflects the instance. 

After evaluating all the operations, Cobweb chooses the operation that yields the maximal average category utility. When it elects to add or merge, the tree is updated to reflect these operations and the entire process repeats recursively at the updated node ($C^1_k$ or the new merged node). When splitting is selected, Cobweb removes the split node, promotes the children, and then recursively repeats its evaluation at the current node. Finally, when Cobweb creates a new node, the process terminates.
Figure~\ref{fig:cobweb_learning} illustrates the process of Cobweb learning a new instance.



After the learning phase and the construction of the tree, Cobweb can apply its learned structure to predict the values of unobserved features of a given instance $\bm{x}^*$ with unknown feature(s). This prediction process is similar to the learning phase: the instance traverses down the tree from the root to a leaf through iterative simulations of insertion into the ``best'' child node within each subtree. This traversal results in a path of category nodes visited, akin to those depicted in Figure~\ref{fig:cobweb_learning}. However, unlike the learning process, none of the concept frequency tables are updated to reflect $\bm{x}^*$ and restructuring operations (merging, splitting, and new) are not considered. Subsequently, Cobweb predicts the unobserved feature values by using a specific node along the categorization path. The analysis of \citeA{corter1992explaining} suggests the \textit{basic-level} node, which holds the highest category utility value, should be utilized for inference. However, recent studies with Cobweb \cite{maclellan2016trestle, maclellan2022efficient, maclellan2022convolutional} have favored using the leaf node, claiming that it often yields superior predictive performance.

Cobweb is an example of a \textit{human-like learning} system, meeting the computational ``gauntlets'' outlined by \citeA{langley2022computational}.
Further, \citeA{fisher1990structure} showed how Cobweb can account for various human concept learning effects, including \textit{basic-level} \cite{murphy1982basic, hoffmann1983objektidentifikation}, \textit{typicality} \cite{rosch1975family}, and \textit{fan} effects \cite{anderson1974retrieval}. Their simulations and analyses predicted human response times in psychological studies using the notion of \textit{category match}. This is a variation of category utility that measures the features present in a given classified observation. 
A notable feature of Cobweb is its direct utilization of hierarchical structure. This distinguishes it from many cognitive science and artificial intelligence models that partition observations into flat clusterings labeled by external categories. This hierarchical approach lets Cobweb capture different categorization effects on basic-level and subordinate concepts: while typical objects tend to be classified into basic-level concepts, atypical objects within a category can be classified into subordinate categories instead because of their low intra-category and high inter-category overlaps.

Despite these desirable features of Cobweb \cite{langley2022computational}, its cognitive plausibility and potential to capture psychological effects within human categorization have been unexplored. Can Cobweb account for the human categorization data that other categorization models in cognitive science handle? Does it exhibit more prototype- or exemplar-like categorization behavior, or is it more of a hybrid model? Do the varying concept levels within the Cobweb tree make Cobweb a more flexible categorization model? This paper addresses these questions through several computational experiments focused on seminal cognitive science studies.

\section{Experiments}

To evaluate Cobweb's alignment to further aspects of human concept learning, we conducted computational experiments using the empirical paradigms developed by \citeA{medin1978context} and \citeA{shepard1961learning}. 

\subsection{Learning and Predicting}

Our experiments utilize the implementation of Cobweb developed by \citeA{maclellan2016trestle}\footnote{The codes for the experiments are available at\\ \url{https://github.com/Teachable-AI-Lab/cobweb-psych}}. 
This implementation employs the \textit{information-theoretic} variant of category utility \cite{corter1992explaining} for learning and prediction. Given a category $c$, its \textit{uncertainty} (or \textit{entropy}) is given by
\begin{equation}
    U(c) = \sum_i P(X_i|c)U(X_i|c)
\end{equation}
where 
\begin{equation}
    U(X_i|c) = -\sum_j P(x_{ij}|c)\log P(x_{ij}|c)
\end{equation}
is the uncertainty of the feature $X_i$ given the concept $c$. Here $P(x_{ij}|c)$ is the probability that feature $X_i$ has value $x_{ij}$ given $c$.
The information-theoretic category utility is then defined as:
\begin{equation}
    CU(c) = P(c)[U(c_p)-U(c)]
\end{equation}
where $c_p$ is the parent concept of $c$. This measure captures the informativeness of the category in terms of the expected reduction in feature value uncertainty given knowledge of the child category label over knowledge of the parent label.

Once the Cobweb tree structure is induced using training stimuli, the categorization of a test stimulus $\bm{x}$ with unobserved feature(s) occurs along a path from its root to a leaf node, defining a concept path. To determine the category for predicting the unobserved feature values, we explore two levels of the hierarchy:

\subsubsection{\texttt{leaf}} Cobweb predicts the unobserved features based on the data stored in the leaf node, which is at the subordinate level. This prediction tends to be deterministic due to the certainty of feature values stored at the lowest level. For instance, consider the leaves depicted in Figure~\ref{fig:cobweb_learning}, where all attributes have specific values with 100\% certainty.

\subsubsection{\texttt{basic}} Cobweb predicts the unobserved features based on the data stored at the basic level, which is the node along the categorization path with the highest category utility. In general, these nodes are more superordinate than leaf nodes.

\subsection{The \citeA{medin1978context} Experiments} \label{subsec:medin}

\subsubsection{Dataset and Original Study}

\citeA{medin1978context} proposed the exemplar (i.e., context) model of classification and evaluated it in an artificial category learning experiment with two sets of 16 stimuli that differ on four binary dimensions, and in particular, \textit{color \{red, blue\}}, \textit{form \{triangle, circle\}}, \textit{size of each component \{large, small\}}, and \textit{number of components \{1, 2\}} for geometric stimuli.
Stimuli 4, 5, 7, 13, and 15 composed the training stimuli for Category A, and stimuli 2, 10, 12, and 14 the training stimuli for Category B. The remaining Stimuli (i.e., 1, 3, 6, 8, 9, 11, 16) were the transfer stimuli. Participants initially learned the nine training stimuli and were provided feedback. After engaging in an interpolated activity, they classified all 16 training and transfer stimuli, this time without feedback. \citeA{medin1978context} compared the predicted probabilities generated by the exemplar model for Stimulus 4 and 7. The purpose was to infer whether people learned more prototype-like or more exemplar-like representations.
A higher predicted Category A probability for Stimulus 4 would suggest a closer alignment with the prototype model due to its greater resemblance to the prototype stimulus for Category A, Stimulus 1 (i.e., (1, 1, 1, 1)). By contrast, a higher predicted probability for Stimulus 7 would indicate greater alignment with exemplar representations as it is more similar to the individual stimuli of Category A than the individual stimuli of Category B. 

\subsubsection{Method and Hypothesis}


The process starts by training Cobweb with the 9 designated training stimuli, then obtaining predicted probabilities for all 16 training and transfer stimuli. We compare these probabilities with the human classification probabilities from the original study using the Pearson correlation coefficient and root mean squared deviation (RMSD) to quantify the relative and absolute fit, respectively.

We derive predictions using two methods, \texttt{leaf} and \texttt{basic}. To ensure the robustness and reliability of experimental outcomes, we conduct the experiments using 40 different random seeds when randomizing the stimuli learning order, and each seed is associated with 5 repeated implementations (so each stimulus is predicted 200 times). This handles stochasticity introduced because, in cases of tied expected category utility, Cobweb randomly selects the best next operation.

We expect Cobweb to exhibit a strong alignment with human data with both prediction methods. Furthermore, we have no \emph{a priori} expectation that Cobweb will strictly adhere to either a prototype or exemplar model, so Stimuli 4 and 7 may exhibit less striking differences in predicted probability than was observed in the human data. In fact, we expect a slightly higher predicted probability for Stimulus 4, which would suggest a more prototype-like categorization by Cobweb. These expectations are rooted in the idea that Cobweb builds up a hierarchical cognitive structure of concepts, generating predictions based on a specific concept node with more or less integrated information.

\subsubsection{Results and Discussions}


\begin{table}[t!]
\small
\centering
\caption{The left panel shows the observed classification probabilities from human subjects in the \citeA{medin1978context} study when using geometric stimuli and the predicted classification probabilities of Cobweb at two levels, \texttt{leaf} and \texttt{basic}, on the stimulus's respective classifications. For instance, the classification probability of Stimulus 4(A) is the probability that Stimulus 4 is classified as Category A, and the one of Stimulus 2(B) is the probability that Stimuli 2 is classified as Category B. To facilitate a direct comparison of the classification probabilities of Stimuli 4(A) and 7(A), we indicate them in shaded rows and denote the stimulus with the greater classification probability using \textbf{bold} text. The right columns show the sample standard deviation of the predicted probability of each stimulus at either the leaf or basic level.} 
\label{tb:medin-probability}
\vskip 0.12in
\begin{tabular}{c|ccc|cc} 
\hline\hline
\multirow{2}{*}{{\bf Stimulus}} & \multicolumn{3}{c|}{{Mean Probability}} & \multicolumn{2}{c}{Sample SD}\\
  & \texttt{\textbf{human}} & \texttt{\textbf{leaf}} & \texttt{\textbf{basic}} & \texttt{\textbf{leaf}} & \texttt{\textbf{basic}} \\
\hline
\multicolumn{6}{c}{Training Stimuli}\\\hline
\rowcolor{lightgray} 4A & 0.780 & 0.735 & \textbf{0.837} & 0.061 & 0.161\\
\rowcolor{lightgray} 7A & \textbf{0.880} & \textbf{0.750} & 0.826 & 0.000 & 0.196\\
15A & 0.810 & 0.750 & 0.893 & 0.000 & 0.048\\
13A & 0.880 & 0.750 & 0.854 & 0.000 & 0.114\\
5A & 0.810 & 0.750 & 0.796 & 0.000 & 0.204\\
12B & 0.840 & 0.695 & 0.664 & 0.154 & 0.300\\
2B & 0.840 & 0.723 & 0.751 & 0.109 & 0.233\\
14B & 0.880 & 0.750 & 0.839 & 0.000 & 0.156\\
10B & 0.970 & 0.750 & 0.867 & 0.000 & 0.078\\\hline
\multicolumn{6}{c}{New Transfer Stimuli}\\\hline
1A & 0.590 & 0.685 & 0.784 & 0.163 & 0.234\\
6A & 0.940 & 0.750 & 0.885 & 0.000 & 0.069\\
9A & 0.500 & 0.350 & 0.220 & 0.206 & 0.239\\
11A & 0.620 & 0.675 & 0.724 & 0.166 & 0.286\\
3B & 0.690 & 0.605 & 0.651 & 0.226 & 0.323\\
8B & 0.660 & 0.650 & 0.721 & 0.201 & 0.282\\
16B & 0.840 & 0.750 & 0.828 & 0.000 & 0.146\\
\hline\hline

\end{tabular}  
\end{table}

\begin{table}[t!]
\small
\centering
\caption{The correlation coefficients and RMSD values between the predicted and observed classification probabilities (shown in Table~\ref{tb:medin-probability}) for the geometric stimuli in the \citeA{medin1978context} experiment.} 
\label{tb:medin-correlation} 
\vskip 0.12in
\begin{tabular}{c|cc} 
\hline\hline
\textbf{Stimuli Set}  &  \texttt{\textbf{leaf}} & \texttt{\textbf{basic}}\\
\hline
Correlation & 0.768 & 0.713 \\
RMSD &  0.166 & 0.130 \\
\hline\hline
\end{tabular}  
\end{table}




The observed and predicted classification probabilities for each stimulus with their sample standard deviations are listed in Table~\ref{tb:medin-probability}. The corresponding correlation coefficients and RMSD values are presented in Table~\ref{tb:medin-correlation}. Considering the predicted probabilities for Stimuli 4 and 7 in Table~\ref{tb:medin-probability}, Cobweb exhibits a slightly higher probability for Stimulus 7 with the \texttt{leaf} level prediction (0.735 vs. 0.750), but a slightly higher probability for Stimulus 4 with the \texttt{basic} level prediction (0.837 vs. 0.826). Although the differences here are not very significant, they are less likely to be affected by the variance of the predicted probabilities given relatively small sample standard deviations for predicted probabilities at both levels with a sample size of 200 each. This pattern shows that Cobweb does not strictly adhere to a prototype- or exemplar-like categorization model paradigm. Instead, it appears to exhibit aspects of both. This capability aligns with the insights from \citeA{fisher1990structure}, particularly the typicality effects observed in Cobweb: the distributed categorization strategy employed by Cobweb allows atypical objects to be categorized into subordinate-level concepts because of low intracategory and high intercategory overlaps.

By examining the correlation scores and the corresponding RMSD values compared with predicted and observed human probabilities, both prediction levels (\texttt{leaf} and \texttt{basic}) employed by Cobweb result in a strong correlation and a modest amount of absolute error with the human data, demonstrating alignment with human concept learning.

\subsection{The \citeA{shepard1961learning} Experiments} \label{subsec:shepard}

\subsubsection{Dataset and Original Study}


In the original \citeA{shepard1961learning} study, there are 8 stimuli and they differ on 3 binary dimensions - \textit{size \{small, large\}}, \textit{color \{white, black\}}, and \textit{form \{square, triangle\}}. Over these, six category structures I-VI are defined, wherein each category A and B span 4 stimuli each. Each structure is defined by a logical rule of increasing complexity distinguishing A from B.
Type I concerns a single diagnostic dimension: the stimuli in each category just differ in color. Type II, the correlated-features task, instantiates the XOR problem along two of the dimensions. Tasks III and V are rule-plus-exception tasks, with the rule leaving an exception item that requires an additional conjunct to represent, and thus presumably additional cognitive processing to learn. Task IV mainly concerns the family resemblance---the prototypes of each category (a large black triangle for Category A and a small white square for Category B) are joined by the stimuli that share two of three features with their prototype. Type VI is arbitrary and the stimuli of each category are special cases sharing no common structure.

\citeA{smith2004category} replicated the original \citeA{shepard1961learning} study using a more comprehensive approach to increase the robustness of the results. Each task type encompassed 6 possible stimuli arrangements (permutations), resulting in 36 distinct tasks. Each human participant engaged in six tasks, each randomly selected from the available permutations within the respective task type. For each task, participants underwent a learning phase lasting 24 blocks (or iterations), on each of which they saw all eight stimuli, equating to $24\times8=192$ trials overall.

In the original study, \citeA{shepard1961learning} concluded that humans do not simply follow behaviorist laws of conditioning and stimulus generalization, and instead ``abstract dimensions [and] then formulate and test rules about how the values on those dimensions combine and interact to determine which classificatory response will be correct'' (p. 33).

\subsubsection{Method and Hypothesis}


In the replication of \citeA{smith2004category}, each task type is instantiated as six tasks (so there are 36 tasks in total), and each task is a permutation of the eight stimuli among two categories such that they satisfy the rule specified by the task type. We ran Cobweb on each of the 36 tasks, repeating each five times with different random seeds.
For each task repetition, after training on all stimuli, Cobweb was used to predict the category of the eight trained stimuli, and we computed an average accuracy score based on these predictions.

We compared observed human and model-predicted accuracies separately on each of the six task types. For each task in its task type, Cobweb was trained across 24 blocks, each running through the randomly ordered 8 stimuli. After each block, the average accuracy over the 8 stimuli was computed. These results are averaged across the six tasks for each type to produce an average learning curve across the 24 blocks. Finally, these model-predicted learning curves are compared to the human learning curves from \citeA{smith2004category}, and correlation coefficients and RMSD values quantify their correspondence. 
Note that \citeA{smith2004category} provide human accuracies for only the odd learning blocks (1, 3, 5, \dots, 19, 21, 23), and so performance on these blocks is the basis of the comparison between the human and Cobweb learning curves.



We expect Cobweb to show strong alignment with human learning across the six task types: to learn the simpler Type I and II structures more rapidly and to a higher accuracy level than the more complex Type V and VI structures. We also conjectured the \texttt{leaf} predictions may be less comparable to the human learning data: Because the leaf nodes contain the homogeneous feature values only, the predictions made by these nodes are always the same. Thus, their predictions are ``overly'' deterministic compared to human predictions, which might make their learning curves artifactually resemble a horizontal line.

\subsubsection{Results and Discussion}

\begin{figure}[t!]
\centering
\includegraphics[width=0.45\textwidth]{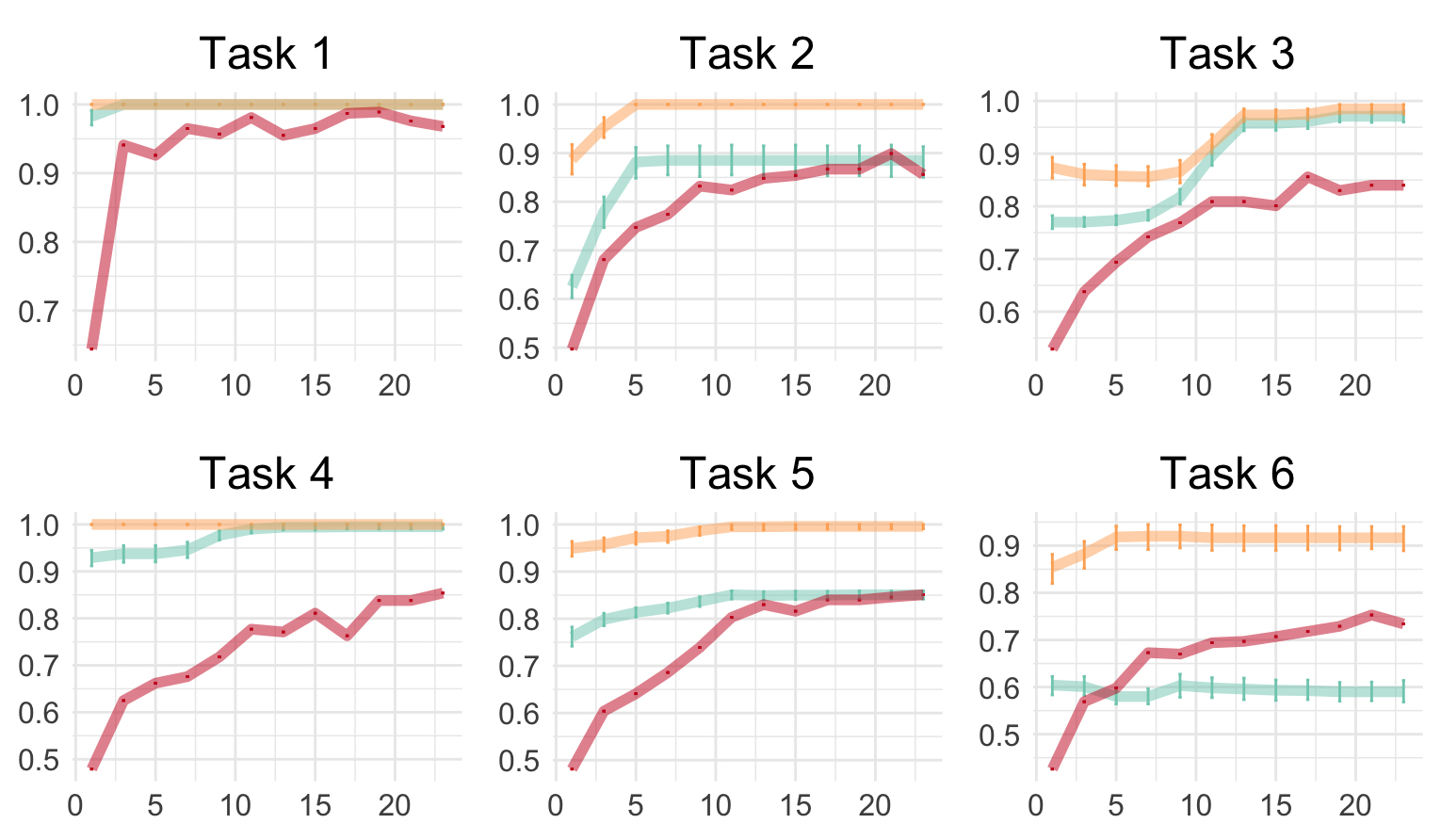}
\caption{The learning curves for human participants (\textbf{\textit{red}}), the \texttt{leaf} prediction level (\textbf{\textit{orange}}), and the \texttt{basic} prediction level (\textbf{\textit{light green}}) across the learning blocks $1-23$. Learning block is on the $x$-axis and (human and model) accuracy on the $y$-axis.}
\label{fig:shepard-exp2}
\end{figure}

\begin{table}[t!]
\small
\centering
\caption{Correlation coefficients and RMSD values with the \citeA{shepard1961learning} category structures. The values are computed by comparing the accuracy scores from human participants of the \citeA{smith2004category} replication and Cobweb's \texttt{leaf} or \texttt{basic} prediction levels, respectively, across the learning blocks $1, 3, ..., 21, 23$. N/A indicates that the correlation coefficient cannot be computed because one compared set of data (the accuracy score set of Cobweb among all 11 training blocks) is constant.}
\label{tb:shepard-exp2} 
\vskip 0.12in
\begin{tabular}{ccccccc} 
\hline\hline
\textbf{Level} & \textbf{I} & \textbf{II} & \textbf{III} & \textbf{IV} & \textbf{V} & \textbf{VI} \\
\hline
\rowcolor{SpringGreen} \multicolumn{7}{c}{Correlation}\\
\texttt{leaf} & N/A & 0.929 & 0.751 & N/A & 0.984 & 0.884 \\
\texttt{basic} & 0.981 & 0.932 & 0.841 & 0.910 & 0.984 & -0.375 \\
\hline
\rowcolor{GreenYellow} \multicolumn{7}{c}{RMSD}\\
\texttt{leaf} & 0.110 & 0.206 & 0.153 & 0.231 & 0.210 & 0.145 \\
\texttt{basic} & 0.105 & 0.075 & 0.112 & 0.198 & 0.089 & 0.230\\
\hline\hline
\end{tabular} 
\end{table}

Figure~\ref{fig:shepard-exp2} shows the learning curves for each of the six task types -- both the observed human data and the two predictor levels of Cobweb. The corresponding correlation coefficients and RMSD values are presented in Table~\ref{tb:shepard-exp2}. Overall, Cobweb shows promising alignment for most task types and for both prediction levels.

In Task I (diagnostic task), both humans and Cobweb learn the categories rapidly and accurately, and the \texttt{leaf} prediction even achieves 100\% accuracy after the first learning block, resulting in a horizontal learning curve across blocks. The \texttt{basic} learning curve has a high correlation with human data ($r = 0.981$). A similar outcome is observed in Task IV (family resemblance). The \texttt{leaf} prediction achieves perfect accuracy, making it challenging to compute its correlation coefficient with human performance. The \texttt{basic} predictions also exhibit a strong correlation with humans ($r = 0.910$).

For Task II (XOR), both prediction levels achieve high correlations with the human data ($r = 0.929$ and $0.932$). For Type V, the correlations are again high and comparable ($r = 0.984$ for both levels). However, for Type III, another rule-plus-exception task type, the correlation coefficients are lower compared to Task V: $r = 0.751$ for \texttt{leaf} and $0.841$ for \texttt{basic}. One possible explanation for this difference is that Task V involves an exception for a single rule, whereas Task III involves an exception for two rules. This more complex scenario results in Cobweb requiring more blocks before a rapid accuracy boost. Finally, in Task VI (no family resemblance), \texttt{leaf} predictions are again strongly aligned with human performance, and overall accuracy remains high at around 0.90 (though it is less than for the other, simpler types). However, \texttt{basic} predictions underperform for Task VI, both in terms of human alignment and overall accuracy. Recall that this is the ``chaotic'' conceptualization, i.e., there is no distinct typical or atypical stimulus for each category. Consequently, neither the basic-level concepts nor the subordinate concepts perform well.

Note that, although Cobweb exhibits promising alignment with the human data at both prediction levels, as shown by correlation coefficients, the RMSD values diverge from 0 across most task types at both levels. Making more accurate absolute predictions is a challenge for the future development of Coweb as a model of human categorization.

Finally, we explored Cobweb's ability to predict the relative difficulty of the six task types I-VI for the humans in the \citeA{shepard1961learning} replication of \citeA{smith2004category}. Table~\ref{tb:shepard-order} provides the observed human data after the first (1) and final (23) learning blocks. Note that there is some stability over learning, with Type I as the easiest and Type VI the hardest in both blocks. The predicted difficulty rankings of the task types for those two blocks is also shown in the table. The alignment is promising, with the basic method ranking on the first learning block and the leaf method ranking on the final learning block agreeing with the human rankings well.

\section{Discussion}

This paper has evaluated the alignment of \textit{Cobweb}, a classical AI model of incremental concept learning, against data from two seminal cognitive science experiments, \citeA{medin1978context} and \citeA{shepard1961learning}. The promising alignment between human performance and Cobweb's predictions demonstrates its viability as a cognitive science model, adding to the evidence provided by an earlier evaluation \cite{fisher1990structure}.

The hierarchical structure of Cobweb enables it to generate predictions at different levels. Here, we derive categorization predictions at two levels: the \texttt{leaf} (i.e., subordinate) level and the \texttt{basic} level. A notable feature is that the flexibility of Cobweb is that it can span the spectrum between prototype-like and exemplar-like representations.  
This flexibility may enable it to account for the transition from prototype representations early in concept acquisition to exemplar representations after extended learning \cite{smith1998prototypes}.

These findings are a first step in demonstrating Cobweb's potential as a model of human categorization. It is important to note that in the experiment by \citeA{medin1978context}, our comparison between Cobweb's predictions and observations was limited to geometric stimuli, whereas the original study covered two sets of stimuli (geometric stimuli and Brunswik faces), both share nominal representations which are simplified and artificially constructed. Indeed, many prior studies comparing exemplar and prototype models have utilized highly simplified perceptual stimuli and artificially designed category structures rather than more natural stimuli and natural category domains. One limitation of this approach is that participants typically have extensive prior experience with the categories being tested, and this learning history is not controlled in experiments \cite{nosofsky2022contrasting}. To bridge this gap and better understand categorization in more natural settings with more complex and high-dimensional stimuli, \citeA{battleday2020capturing} employed various machine learning methods on a large behavior dataset featuring natural images. Moving forward, our experiments can be extended using Cobweb/4V \cite{barari2024avoiding}, a derivative of Cobweb that incorporates image representations instead of low-dimensional artificial ones. This could let Cobweb account for categorization effects in studies with natural images.

\begin{table}[t!]
\small
\centering
\caption{Comparison of the relative difficulty of the six task types I-VI after the first (1) and last (2) learning blocks by both humans \cite{smith2004category} and Cobweb (leaf, basic-level nodes). Human rankings are highlighted with shaded rows, with matching task types indicated in {\bf bold} text at corresponding ranking positions. The Spearman's rank correlation coefficients $\rho$ between human and predicted rankings are in the right column, where tied ranks are considered.}
\label{tb:shepard-order} 
\vskip 0.12in
\begin{tabular}{ccccccc|c} 
\hline\hline
\textbf{Ranking} & 1 & 2 & 3 & 4 & 5 & 6 & $\rho$\\
\hline
\multicolumn{8}{c}{Block 1}\\
\rowcolor{lightgray} Observed & I & III & II & V & IV & VI & \\
\texttt{leaf} & {\bf I} & IV & V & II & III & {\bf VI} & 0.386\\
\texttt{basic} & {\bf I} & IV & III & {\bf V} & II & {\bf VI} & 0.829\\
\hline
 \multicolumn{7}{c}{Block 23}\\
\rowcolor{lightgray} Observed & I & II & IV & V & III & VI & \\
\texttt{leaf} & {\bf I} & {\bf II} & {\bf IV} & {\bf V} & {\bf III} & {\bf VI} & 0.857 \\
\texttt{basic} & {\bf I} & IV & III & II & V & {\bf VI} & 0.486\\
\hline\hline
\end{tabular} 
\end{table}

Future research should also expanded the evaluation of Cobweb to other important findings on human categorization. Building on \citeA{shepard1961learning}, can Cobweb also account for studies of structured concepts \cite{feldman2000minimization, feldman2003catalog, hayes1977concept}? Building on \citeA{medin1978context}, can it account for studies of linear separability \cite{medin1981linear, levering2020revisiting} and correlated features \cite{malt1984correlated, medin1982correlated}? And finally, what is its relationship to other ``hybrid'' models like RULEX \cite{nosofsky1994rule} and SUSTAIN \cite{love2004sustain}?

In conclusion, we provide preliminary evidence that Cobweb can flexibly model human category learning---exhibiting both exemplar- and prototype-like behavior. We look forward to evaluating it across more study data and developing it into a robust model of human category learning. 


\bibliographystyle{apacite}

\setlength{\bibleftmargin}{.125in}
\setlength{\bibindent}{-\bibleftmargin}

\bibliography{CogSci_Template}

\end{document}